\documentclass[letterpaper, 10 pt, conference]{ieeeconf}

\IEEEoverridecommandlockouts
\usepackage[binary-units]{siunitx}
\usepackage{graphicx}

\newcommand{\res}[2]{\SI{#1}{}\,x\,\SI{#2}{}}

\title{\LARGE \bf
Are we ready for beyond-application high-volume data?\\The Reeds robot perception benchmark dataset*
}

\author{Ola Benderius$^{1}$ and Christian Berger$^{2}$ and Krister Blanch$^{1}$
\thanks{*This work was supported by The Swedish Transport Administration}
\thanks{$^{1}$Ola Benderius and Krister Blanch are with Department of Mechanics and Maritime Sciences, Chalmers University of Technology, Sweden
        {\tt\small \{ola.benderius,krister.blanch\}@chalmers.se}}%
\thanks{$^{2}$Christian Berger is with the Department of Computer Science and Engineering, University of Gothenburg, Sweden
        {\tt\small christian.berger@gu.se}}%
}

\begin{document}

\maketitle
\thispagestyle{empty}
\pagestyle{empty}

\begin{abstract}


This paper presents a dataset, called Reeds, for research on robot perception algorithms. The dataset aims to provide demanding benchmark opportunities for algorithms, rather than providing an environment for testing application-specific solutions. A boat was selected as a logging platform in order to provide highly dynamic kinematics. The sensor package includes six high-performance vision sensors, two long-range lidars, radar, as well as GNSS and an IMU. The spatiotemporal resolution of sensors were maximized in order to provide large variations and flexibility in the data, offering evaluation at a large number of different resolution presets based on the resolution found in other datasets. Reeds also provides means of a fair and reproducible comparison of algorithms, by running all evaluations on a common server backend. As the dataset contains massive-scale data, the evaluation principle also serves as a way to avoid moving data unnecessarily.

It was also found that naive evaluation of algorithms, where each evaluation is computed sequentially, was not practical as the fetch and decode task of each frame would not scale well. Instead, each frame is only decoded once and then fed to all algorithms in parallel, including for GPU-based algorithms.

\end{abstract}

\section{INTRODUCTION}

The need to benchmark and compare the performance of robot perception algorithms is crucial for the development of autonomous robots acting in complex environments. For this purpose numerous datasets have been introduced in the past, starting more than two decades ago with relatively small and specialized sets (e.g.~\cite{liu2007performance}\cite{dollar2009pedestrian}\cite{brostow2009semantic}), and then about ten years ago continuing with more general and mature sets. One such example is the well-known Kitti vision benchmark suite~\cite{geiger2012we} specifically targeting the development of perception algorithms for autonomous road vehicles.  In a recent overview of open datasets it was found that at least thirty-seven sets using road vehicles were published between 2007 and 2018~\cite{kang2019test}. Interestingly from the summary, from a meta-analysis of the paper, two clear technical trends can be seen based on the release dates of the surveyed datasets, one around 2009 corresponding to the increased development and adoption for \textit{advanced driver-assistance systems}~(ADAS) for road vehicles, and one in 2016 corresponding to the accelerated race around \textit{autonomous drive}~(AD). Furthermore, anecdotally, one can also see from the scope of the datasets, which aim at supporting the development of ADAS, focus is primarily given to formally non-complex applications such as detection of discrete objects (e.g., pedestrians, signs, and lane markings) and semantic segmentation, while datasets targeting AD, formally complex applications are targeted. Typical areas within perception research include (i)~stereo vision and depth perception, (ii)~optic and scene flow, (iii)~odometry and simultaneous localization and mapping, (iv)~object classification and tracking, (v)~semantic segmentation, and (vi)~scene and agent prediction. Since 2019, there has been some noteworthy other datasets, such as the H3D dataset~\cite{patil2019h3d} and the Waymo dataset~\cite{sun2020scalability}. There are also datasets created using unmanned aerial drones~(UAVs) for similar purposes, such as UAV123~\cite{mueller2016benchmark} and MOR-UAV~\cite{mandal2020mor}. However, in general, datasets are still very limited from the environments where they are collected, primarily either from on-road scenarios using conventional passenger cars with significant limitations on kinematic capabilities, or from flying drones with significant limitations on the equipped sensor package. Arguably, to fully test the accuracy and performance of robot perception, a state-of-the-art dataset should enable both: a kinematically demanding agent, as given by a flying platform, as well as a sensor package providing high-volume and high-quality data of diverse scenarios, which brings high demands on the physical and electrical infrastructure of the agent. In this work, it was concluded that a marine vehicle would be the best agent to fulfill these requirements, as weather conditions directly impact the kinematics, and that it can be sufficiently equipped. In addition, marine environments also provide both sparse and dense visual and geometric sensor feedback, a large verity of discreet man-made and biological objects, and predictable behavioral patterns---which are all relevant for the typical types of algorithms. Uniquely, it also provides morphological geometry as given by the surrounding water, giving further opportunities for algorithms. It can be then argued that a perceptually demanding dataset is preferred compared to an application-centric one, to better support basic research in perception algorithm development and the corresponding benchmarking. In this paper, a newly published dataset, referred to as Reeds, fulfilling the above requirements and goals are described.

What has been missing from relevant benchmark suites so far are means of fair and reproducible benchmarks of algorithm computational performance, as recently clearly pointed out in detail~\cite{nguyen2019systematic}. It was also found that benchmarks towards open datasets were primarily focusing on \emph{accuracy} rather than run-time performance such as for example evaluation of formal real-time capabilities (i.e., suitability to run in embedded robotic systems). Furthermore, it was found that the reported results from a few research studies could not be confirmed due to lack of conserved and fully linked software, missing documentation, or missing source code to enable full replicability and to foster further research and competition within a certain research direction. In addition, since benchmarks are typically computed only when the publication in question is submitted, or possibly in some cases occasionally by the database providers for the purpose of updating the leaderboards over time, the specific numbers can rarely be trusted to be fair, for instance, due to differences in computational platforms. For the dataset that is proposed here, benchmark results should always be computed and presented fairly, using automated containerized evaluation as proposed previously~\cite{nguyen2019systematic}. This objective also coincides with another database objective, to maximize data volume and quality so that algorithms also can compete on various preset data spatiotemporal resolution profiles rather than on a single setting. As a design consequence from fair comparison and data volumes that cannot easily be downloaded, any researcher or developer needs to register and submit the algorithm into the Reeds cloud environment in order to compete with new algorithms towards the Reeds database. This is either as a link towards an open-source repository that is compiled on the fly, or a binary for every new comparison towards other algorithms.

Due to the goal of high-volume data (i.e.~larger than what is practical to download and use locally), with the implication that algorithms need to be evaluated close to the data, an important technical problem of this research study was to find ways to evaluate algorithms towards the data as efficiently as possible. The research questions of this work is therefore:
\begin{itemize}
    \item[RQ1] What is the design of cloud environment focused on massive-scale datasets to allow for robot perception algorithm benchmarking?
    \item[RQ2] How does the evaluation time of a large set of perception algorithms scale when used on large data sizes (including several preset down-scaled versions), and how can the evaluation time be minimized so that full re-evaluation towards all data is feasible?
\end{itemize}

\section{THE DATASET}

\begin{figure}[t]
    \centering
    \includegraphics[width=\linewidth]{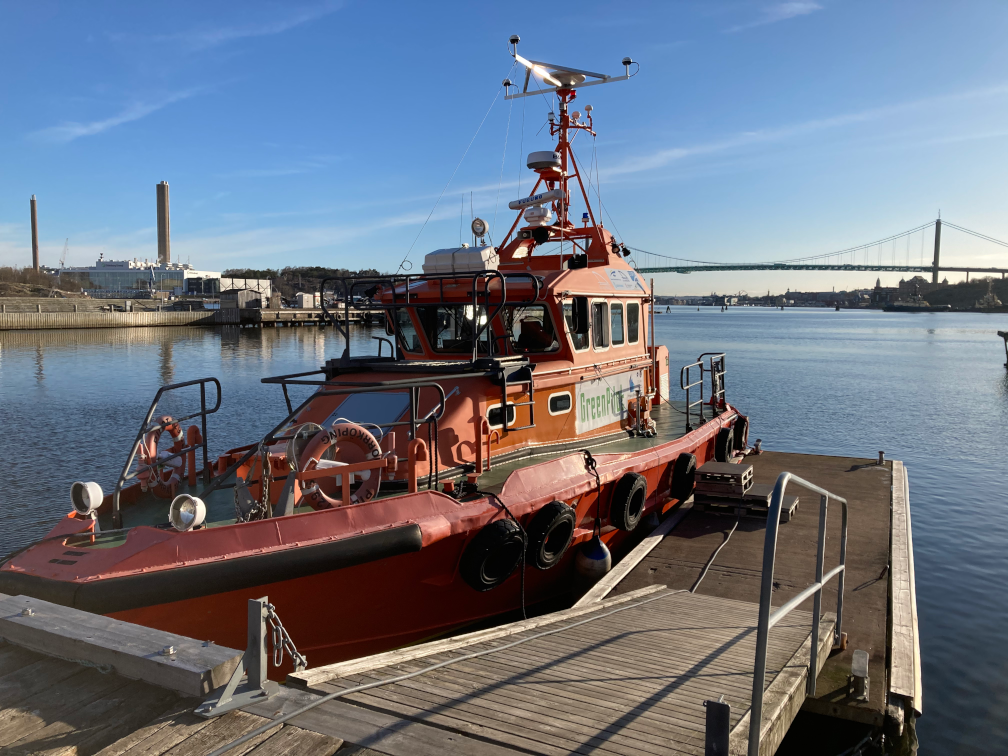}
    \caption{The boat used to carry the sensor platform and logging equipment.}
    \label{fig:boat}
\end{figure}

\begin{figure}[t]
    \centering
    \includegraphics[width=\linewidth]{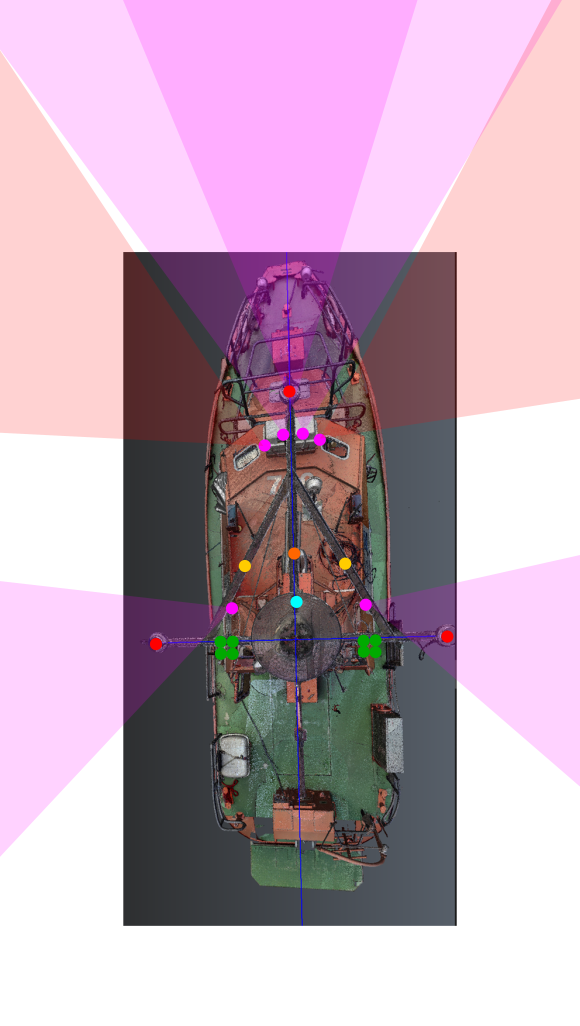}
    \caption{An overview of the sensor platform. The pink dots correspond to the high-performance vision sensors, where the pink areas correspond to monochrome field of view (\SI{43.2}{\degree}) and the orange areas to three-channel (RGB) field of view (\SI{43.2}{\degree}). Note that good stereo coverage is given towards the front. Furthermore, the red dots corresponds to the GNSS antennas, the orange to the IMU, the teal to the \SI{360}{\degree}~radar, the green to the \SI{360}{\degree}~documentation camera system, and the yellow to the lidar sensors.}
    \label{fig:sensors}
\end{figure}

The platform for data collection is a \SI{13}{\meter} retired pilot boat, as depicted in Fig.~\ref{fig:boat}. The basic sensor platform, as described in Table~\ref{tab:sensors} and Fig.~\ref{fig:sensors}, consists of a GNSS system with three antennas and real-time kinematic positioning capabilities and a fibre optic gyro IMU system, four single-channel high-performance vision sensors, two three-channel high-performance vision sensors, a 360\si{\degree}~documentation camera system, two high-performance laser range scanners, a conventional marine 360\si{\degree}~spinning radar sensor, a weather and barometric pressure sensor, automatic identification system~(AIS), and various engine and vehicle state sensors. The GNSS sensor and the IMU jointly provides ground truth positioning. The sensor package will be kept as constant as possible for all data collection runs, continuing for three years, and each run will include separate calibration data for all sensors. Possibly, for specialized use cases, additional sensors might be added to future data collection runs.

A three-dimensional scan was also made of the boat, resulting in a detailed three-dimensional model to be used as basis for further sensor calibration and simulation. The full model, as illustrated in Fig.~\ref{fig:sensors}, is published as part of the dataset, and all agent-specific local coordinates, such as sensor mount positions, are provided in the model coordinate system.

The logging equipment consists of two data center servers, each equipped with: Two AMD EPYC 7352 CPUs (four in total), \SI{128}{\giga\byte} memory, four \SI{16}{\tera\byte} Seagate Exos SATA disk drives, one \SI{15.36}{\tera\byte} Micron 9300 NVMe U.2 SSD (\SI{3.5}{\giga\byte\per\second} write), one \SI{2}{\tera\byte} Samsung 980 PRO NVMe M.2 SSD (\SI{5.1}{\giga\byte\per\second} write), one Nvidia GeForce RTX 2080, and one Nvidia Quadro RTX 4000. Each server is connected to three Flir cameras and one Velodyne lidar via dedicated 10Gb Ethernet links, and the rest of the sensors are connected via a switched network. The IMU is connected via RS-485 through a dedicated computational unit. All computational nodes run Linux kernel 5.10 with PREEMPT\_RT activated. The system is connected to the Internet using a 4G link. All computational nodes are time synchronized using PTP through a common global clock synced through GPS, except for the Axis system which is synchronised using local NTP.

\begin{table}[h]
    \caption{An overview of installed sensors}
    \label{tab:sensors}
    \centering
    \begin{tabular}{lcc}
        \hline
        \hline
        Type & Qty. & Sensor \\
        \hline
        IMU & 1 & KVH P-1775 \\
        GNSS & 1 (3 ant.) & Anavs MSRTK \\
        1-channel vision & 4 & Flir ORX-10G-71S7M-C \\
         &  & EO 16mm f/4 1" HPr lens (\SI{43.2}{\degree}) \\
        3-channel vision & 2 & Flir ORX-10G-71S7C-C \\
         &  & EO 16mm f/4 1" HPr lens (\SI{43.2}{\degree}) \\
        360\si{\degree} doc. cam. & 2 (8 cam.) & Axis F44 \\
        lidar & 2 & Velodyne Alpha prime \\
        360\si{\degree} radar & 1 & Simrad Halo 20+ \\
        weather sensor & 1 & Airmar 220WXH \\
        \hline
        \hline
    \end{tabular}
\end{table}

\subsection{Data logistics}

The largest data volume is generated by the six high-performance cameras from the Oryx ORX-10G-71S7-series. Each monochrome camera is able to generate up to \SI{0.765}{\giga\byte\per\second} (\res{3208}{2200} maximum resolution in 10-bit depth at 91 fps); the color variant is delivering about the same rate according to the specifications using the BayerRG10p pattern and about \SI{5}{\percent} more using a YUV444 pattern. Each frame is fed through the Nvidia Quadro card for lossless compression; initial experiments with the x265 video compression codec for lossless compression indicate some space savings of around \SI{75}{\percent} for 4K frames under laboratory conditions and the gains for actual recordings may likely differ depending on, for instance, scenery and light conditions. As a comparison, the second largest producer of data are the lidars, each writing 4\,800\,000~points per second corresponding to about \SI{75}{\mega\byte\per\second}. Each of the two servers connects to three cameras, where recordings are written to the \SI{15.36}{\tera\byte} SSD drives. Therefore, the total time that the full logging system can be active is around \SI{76}{\minute}. After such logging run, the SSD can then be dumped onto an SATA disk in about \SI{25}{\hour} for further post-processing into the cloud service backend. Note that a full trip of data correspond to two such disks, resulting in about \SI{30}{\tera\byte} of data.

\subsection{Sensor calibration}

For the cameras, three types of calibration is carried out: Vignette and exposure calibration~\cite{goldman-iccv05-vignetting}, with known response curve, intrinsic calibration, and extrinsic calibration. The extrinsic calibration is carried out before each data collection run, and is automatically done by a sensor fusion approach between cameras, lidars, and the radar towards a \SI{2.4x0.9}{\meter} checkerboard target with a known WGS84 position and heading. The calibration requires a few minutes of measurements, and the process is simplified by large motions in roll and in the z-axis. The position of the three GNSS onboard antennas and the IMU are automatically determined from the measured lidar point cloud motion. The method used to solve the automated calibration task is a genetic algorithm. All calibration values are appended to each data log for later use.

\subsection{Annotation and privacy}

Automatic annotation is employed for ships, using the AIS and lidar sensors. Furthermore, included as part of the Reeds web environment, there is an annotation tool showing objects in all camera views, lidar, and radar, where users are asked to mark objects of different classes using 3D bounding boxes. So far, there are object classes for boats such as leisure sailing boats, leisure motor boats, large passenger ferries, commuting ferries, road ferries, tug boats, cargo ships, tankers ships, but also other objects such as waterway signs, humans in water, humans out of water, birds, bridges, buildings, passenger cars, trucks, and bicyclists. The annotation is typically done at \SI{1}{\hertz}, even if data technically is stored at much higher frequencies. Further object classes are envisioned to be added based in discussion with dataset users.

There is also automated anonymization in place, divided into the two classes: \textit{Strict} as for example related to AIS, where the identification, call sign, name, destination, and estimated time of arrival of other ships are never stored into logs, and \textit{relaxed} where already annotated objects such as boats and persons are blurred in any previewed material in the web environment. There are also some demonstration sets that can be freely downloaded, and these are always checked for any sensitive data before being made downloadable. Note that data in general can never be downloaded, and non-annotated data can only be viewed in the web-browser by persons agreeing to handle the sensitive data with special care. Furthermore, any image that is viewed to an annotator is watermarked with that persons identification, with the instruction that it should never leave the web interface.

\subsection{Data collection plan}

The data collection runs are all planned for the areas around Göteborg, Sweden, both in the river Göta älv close to busy water close to the city center, but also inland around Hisingen and towards the lake of Värnen. In addition, sets of data will be recorded close to coastal regions, in the archipelago, and out on open sea. In all of these, busy areas will be favoured where commercial shipping and other human activities are clearly visible. The purpose of the selected environments is to actively allow for the various typical use cases as mentioned above. If some of the use cases turn out to be underrepresented in data, then the list of geographical routes will be revised.

Each run will be close to the maximum data volume limit, about \SI{76}{\minute}, and they will always start and end at the same geographical position, and in some cases they will pass the same position several times. The reason for this is to support the perceptual problem of so called \textit{loop closing}, relevant for algorithmic sense of direction.

\section{AUTOMATED EVALUATION}

A key feature of the Reeds dataset infrastructure is the automated evaluation of algorithms. Having this in place allows fair reproducible benchmarks of previous and future algorithms. Furthermore, it removes the need of downloading vast amounts of data, but rather move comparatively smaller software executables or source code. However, by also allowing users to utilize the backend infrastructure for evaluation, even if restricted to the officially approved leaderboard candidates, the computational needs on the evaluation servers are significant. Not only restricted to CPU time, but also GPU and wear on disk drives. Therefore, as the most naive way of evaluating algorithms towards the data does not scale well with an increasing number of algorithms, a well designed evaluation procedure was employed.

\subsection{Design}

The overall design of the evaluation infrastructure was made within a focus group interview between five different test projects using early snippets of the dataset. Each project was selected for the purpose of having different requirements from the dataset and the evaluation of the resulting algorithms. The test projects included: Motion estimation and 3D-reconstruction using mono and stereo cameras, object detection and classification, rain-drop removal from video feeds, simultaneous localization and mapping (SLAM) using lidar, and radar-based estimation of motion.

\begin{figure}[t]
    \centering
    \includegraphics[width=\linewidth]{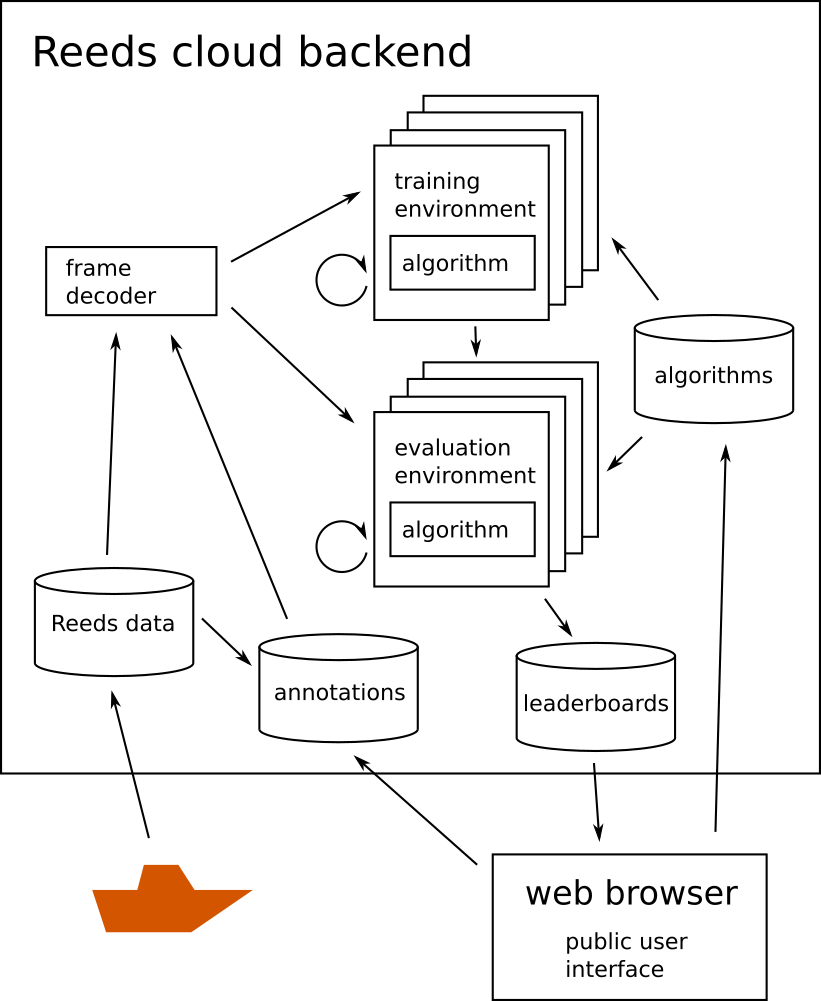}
    \caption{The Reeds automated benchmark environment. Data is transferred from the boat using hard drives. When injected into the backend, automated annotation is started using AIS, lidar, and camera data. Manual annotation is also provided by the integrated web tool, where camera, lidar, and radar data is presented to external users. Algorithms can be uploaded by external researchers and developers. When a new algorithm, a new annotation set, or a new dataset was added, then an automated evaluation is initiated and the results are registered on the leaderboards.}
    \label{fig:benchmark}
\end{figure}

In a first set of meetings, the authors of this paper individually met the principal investigator for each project to discuss their evaluation needs. Then, towards the end of the projects, a joint meeting was organized where the group combined all requirements into a joint proposal that would be able to cater to the needs of each project. Finally, some adjustments to the design were made before reaching the results as shown in Fig.~\ref{fig:benchmark}.

In the design, a common interface towards the dataset is given. Using this interface, data can be consumed by the algorithm in the Reeds cloud environment. The basic method for data access is through shared memory, on the CPU side implemented as POSIX and on the GPU using the \textit{external memory} feature, using VK\_KHR\_external\_memory from Vulkan (a standardized multi-vendor API for GPU access). In order to scale well with an increasing number of algorithms, the data is fed to algorithms in parallel during evaluation. For example, when working with video feeds the involved steps are as follows: (i)~All video frames belonging to a time slice are read from disk and decoded using an Nvidia GPU, (ii)~the resolution of the images and frame rate are stored into shared memory, (iii)~the resulting images are copied to each GPU in the cluster using Vulkan and a handle to the memory is stored in shared memory, (iv)~each algorithm is using the data from the time slice and releases the shared memory when done, (v)~the output from each algorithm is compared towards the ground truth data, and (iv)~when the entire collection run is consumed, the resolution (spatial and temporal) is reduced before starting over at step (ii). The accuracy and precision as well as expectation times of the algorithm is then summarized and automatically reported towards the public leaderboards.

For evaluation, the preset resolution and frame rate profiles were based on other existing datasets. To find these, all the datasets listed by Kang \textit{et al.}~\cite{kang2019test} were reviewed together with common datasets for UAVs, and relevant settings were extraction. It was found that the two common frame rates used are \SI{30}{\hertz} and \SI{10}{\hertz}, and resolutions are typically around \res{1920}{1080} or \res{1280}{720}. An exception is the Apollo dataset which includes videos in the resolution of \res{3384}{2710} at \SI{30}{\hertz}. Furthermore, since the Kitti dataset is still considered very important for evaluations, it was decided to also include the rather unusual resolution of \res{1382}{512} in Reeds as well. A list of all presets are found in Table.~\ref{tab:presets}. In cases where the images could not be scaled with preserved ratio, the overflow was cropped, to avoid any stretching. The cropping is always done relative to the center of the image.

\begin{table}[h]
    \caption{The resolution presets used in the online Reeds evaluation}
    \label{tab:presets}
    \centering
    \begin{tabular}{ccc}
        \hline
        \hline
        Width & Height & Rate  \\
        \hline
        3208 & 2200 & 91/40 \\
        1920 & 1280 & 91/40 \\
        1382 & 512 & 91/40 \\
        1280 & 720 & 91/40 \\
        \hline
        3208 & 2200 & 30 \\
        1920 & 1280 & 30 \\
        1382 & 512 & 30 \\
        1280 & 720 & 30 \\
        \hline
        3208 & 2200 & 10 \\
        1920 & 1280 & 10 \\
        1382 & 512 & 10 \\
        1280 & 720 & 10 \\
        \hline
        \hline
    \end{tabular}
\end{table}

The benefit of the proposed evaluation procedure is that each set of frames are only fetched from disk and decoded once for all algorithms and all down-scaled resolutions, resulting in $n$ fetch and decode operations where $n$ is the number of frames. With the naive method of evaluation, this step would have been repeated for each frame, algorithm, and down-scaled resolution, resulting in $n \times m \times p$ where $m$ is the number of algorithms and $p$ is the number of preset resolutions to evaluate. As $n$ corresponds to the fetching and decoding of about \SI{30}{\tera\byte} of data, and with $p=12$ and a reasonable value $m=100$ (in fact, this could potentially grow to a few hundred over time), the parallel evaluation is expected to save processing worth \SI{36000}{\tera\byte} of data \emph{per data log} by reading the \SI{30}{\tera\byte} only once. The drawback of the parallel evaluation is that each algorithm needs to wait for all other algorithms, per frame.

\section{CONCLUSION}




This paper presents the dataset Reeds, a novel dataset for research and development of robot perception algorithms. The design goal of the dataset was to provide the most demanding dataset for perception algorithm benchmarking, both in terms of the involved vehicle motions and the amount of high quality data. The logging platform consists of an instrumented \SI{13}{\meter} boat with six high-performance vision sensors, two lidars, a \SI{360}{\degree} radar, a \SI{360}{\degree} documentation camera system, and a three-antenna GNSS system as well as a fibre optic gyro IMU used for ground truth measurements. All sensors are calibrated into a single vehicle frame.

In order to offer more fair benchmarking of algorithms, and to avoid moving large amounts of data, the Reeds dataset comes with a evaluation backend running in the cloud. External researchers uploads their algorithms as source code or as binaries, and the evaluation and comparison towards other algorithms is then carried out automatically. The data collection routes were planned for the purpose of supporting stereo vision and depth perception, optic and scene flow, odometry and simultaneous localization and mapping, object classification and tracking, semantic segmentation, and scene and agent prediction. The automated evaluation measures both algorithm accuracy towards annotated data or ground truth positioning, but also per-frame execution time and feasibility towards formal real-time, as standard deviation of per-frame execution time. In addition, each algorithm is also tested on 12~preset combinations of resolutions and frame rates as individually found in other datasets. From this design, Reeds can be viewed as a superset, as it can mimic the data properties other datasets.

Due to the massive-scale data provided by Reeds, and the automated evaluation involving multiple scaled-down versions of the same data, it was found that naive sequential evaluation is not feasible. Instead, a parallel evaluation pattern was employed, where all algorithms and all presets are evaluated in parallel. This is solved by using shared memory for CPU-based algorithms, and the external memory feature of GPUs for GPU-based algorithms. Using this approach, the fetch and decode of data could be lowered to a fraction of what was done in the naive sequential case.

\addtolength{\textheight}{-19.5cm}

\section*{ACKNOWLEDGMENT}

The authors would like to thank Robert Rylander, Ted Sjöblom, and Joakim Lundman at RISE for their support in building the research platform. They would also like to thank Anna Petersson, Varun Hegde, Athanasios Rofalis, Jiraporn Sophonpattanakit, Liangyu Wang, and Iván Garcia Daza for testing and evaluating the dataset.

\bibliographystyle{IEEEtran}
\bibliography{IEEEabrv, refs}

\end{document}